\pdfoutput=1
\documentclass[12pt,a4paper]{article}
\usepackage[T1]{fontenc}
\usepackage{lmodern}
\usepackage[english]{babel}
\usepackage{graphicx}
\usepackage{caption}
\usepackage{amsmath}
\usepackage[toc]{appendix}
\usepackage{amsfonts}
\usepackage{natbib}
\usepackage{xcolor} 
\usepackage{booktabs}
\usepackage{colortbl}
\usepackage{array}
\usepackage{graphicx} 
\usepackage{makecell}
\usepackage{caption}
\usepackage[colorlinks, citecolor=blue]{hyperref}    
\usepackage{authblk} 
\usepackage{listings}
\usepackage{xcolor}
\usepackage{subcaption}
\usepackage{tcolorbox}
\usepackage{multirow}
\usepackage{tcolorbox}

\title{\texttt{Dspy}-based Neural-Symbolic Pipeline to Enhance Spatial Reasoning in LLMs}
\date{\vspace{-2ex}}
\author[1]{\footnotesize Rong Wang \thanks{Correspondence to R. W., \url{Email: rongw.de@gmail.com}}}
\author[2]{\footnotesize Kun Sun}
\author[3]{\footnotesize Jonas Kuhn}

\affil[1,3]{\footnotesize The Institute of Natural Language Processing, Stuttgart University, Stuttgart, Germany}
\affil[2]{\footnotesize Tübingen, Germany}
\begin{document}

	\maketitle
		
	\begin{abstract}
Large Language Models (LLMs) have demonstrated remarkable capabilities across various tasks, yet they often struggle with spatial reasoning. This paper presents a novel neural-symbolic framework that enhances LLMs' spatial reasoning abilities through iterative feedback between LLMs and Answer Set Programming (ASP). We evaluate our approach on two benchmark datasets: StepGame and SparQA, implementing three distinct strategies: (1) direct prompting baseline, (2) Facts+Rules prompting, and (3) DSPy-based LLM+ASP pipeline with iterative refinement. Our experimental results demonstrate that the LLM+ASP pipeline significantly outperforms baseline methods, achieving an average 82\% accuracy on StepGame and 69\% on SparQA, marking improvements of 40-50\% and 8-15\% respectively over direct prompting. The success stems from three key innovations: (1) effective separation of semantic parsing and logical reasoning through a modular pipeline, (2) iterative feedback mechanism between LLMs and ASP solvers that improves program executability rate, and (3) robust error handling that addresses parsing, grounding, and solving failures. Additionally, we propose Facts+Rules as a lightweight alternative that achieves comparable performance on complex SparQA dataset, while reducing computational overhead.Our analysis across different LLM architectures (Deepseek, Llama3-70B, GPT-4.0 mini) demonstrates the framework's generalizability and provides insights into the trade-offs between implementation complexity and reasoning capability, contributing to the development of more interpretable and reliable AI systems.
	\end{abstract}


\newpage
	
	\section{Introduction}

	Large Language Models (LLMs) are known for their impressive performance across a range of tasks, demonstrating certain commonsense reasoning abilities. However, since LLMs are trained to predict subsequent words in a sequence, they seem to lack sufficient grounding to excel at tasks requiring spatial, physical, and embodied reasoning.
    Spatial reasoning, the ability to understand and manipulate relationships between objects in two- and three-dimensional spaces, represents a crucial component of artificial intelligence systems, enabling practical applications in robotics, navigation, and physical task planning. Recent studies (\citealp{bang2023multitask}; \citealp{cohn2023evaluation}) highlighted the limitations of models like \texttt{ChatGPT} in deductive logical reasoning, spatial reasoning, and non-textual semantic reasoning, underlining the need for further improvements in spatial reasoning.

	Efforts to improve logical reasoning capability have focused on advanced prompting techniques. However, these methods have demonstrated notable limitations, particularly on challenging datasets like \texttt{StepGame} \citep{shi2022stepgame} and \texttt{SparQA} \citep{mirzaee2022transfer}, which often require multi-step planning and complex natural language understanding. In these scenarios, LLMs often struggle with maintaining coherence, frequently hallucinating or losing sight of the original objectives, resulting in inaccurate and unreliable outputs. More recent work demonstrates that augmenting large language models (LLMs) with external tools for arithmetic, navigation, and knowledge lookups improves performance \citep{fang2024large}. Notably, neural-symbolic approaches, where LLMs extract facts while external symbolic solvers handle reasoning, show significant promise in enhancing logical inferenceg) \citep{yang2023coupling}. Nevertheless, existing neural-symbolic methods face limitations in generalization, scalability, and comprehensive evaluation. (Yu et al, 2021) Neural-symbolic systems have been applied successfully in visual question answering and robotics \citep{yang2023coupling}.These approaches often fail to fully harness LLMs' potential, particularly by omitting feedback loops between multiple LLMs and symbolic modules, limiting performance gains.
    
    To overcome these challenges, this paper proposes a novel neural-symbolic framework that systematically integrates symbolic reasoning components with LLMs. Our approach leverages strategic prompting, feedback loops, and Answer Set Programming (ASP)-based verification, creating a robust pipeline that improves spatial reasoning across different LLM architectures. We evaluate this framework on two benchmark datasets, By integrating feedback loops and verification module, our methodology demonstrates strong generalizability when tackling complex spatial reasoning tasks, showing significant performance gains, with average accuracy improvements of 40\% on StepGame and 20\% on SparQA. These outcomes underscore the effectiveness of neural-symbolic integration in enhancing spatial reasoning and emphasize the importance of task-specific implementation strategies. 
    
	To address these limitations, we propose a pipeline that effectively enhances LLMs' spatial reasoning capabilities. Our approach combines strategic prompting with symbolic reasoning to create a robust framework that significantly improves performance of spatial reasoning across different LLM architectures. Specifically, building on these insights, we propose a novel neural-symbolic pipeline that integrates LLMs with ASP, aimed at enhancing the LLMs’ capabilities of spatial reasoning in \texttt{SparQA} and \texttt{StepGame} datasets. We investigate the potential benefits of integrating symbolic reasoning components into LLMs to further boost their spatial reasoning capabilities. Within the broader field of neural-symbolic AI, the combination of LLMs as parsers with ASP solvers has emerged as a particularly effective approach for complex reasoning tasks. Additionally, the present study is one of the pioneering projects which uses DSPy \citep{khattab2023dspy} to pipeline and program LLM. 

    In the broader field of neural-symbolic AI, combining large language models (LLMs) with symbolic reasoning systems like Prolog has proven successful for tackling complex reasoning tasks \citep{hamilton2022neural,besold2021neural}.Our study advances this field by demonstrating how incorporating symbolic reasoning modules can enhance LLMs’ spatial reasoning capabilities. Specifically, we introduce a modular pipeline using DSPy, enabling seamless interaction between LLMs and symbolic system solvers. Our pipeline not only improves spatial reasoning but also shows promise in broader domains requiring structured logic, thereby contributing to both the theoretical foundations and practical applications of neural-symbolic integration.

	\section{Preliminaries}
	
\subsection{Prompting LLMs for spatial reasoning}

Research on spatial reasoning spans both visual and textual domains, each with its own challenges. In Visual Spatial Reasoning (VSR), even advanced models like CLIP (Radford et al., 2021) face limitations in comprehending complex spatial relationships. In the textual domain, spatial reasoning is further complicated by linguistic ambiguity.
    
Prompting strategies have played a pivotal role in improving LLMs' reasoning capabilities. Approaches such as \textit{chain of thought} (\texttt{CoT}) (\citealp{Wei2022Chain}; \citealp{Chu2023A}), \textit{few-shot prompting} \citep{Schick2021True}, and \textit{least-to-most prompting} \citep{Zhou2022Least-to-Most} enable LLMs to tackle complex problems by breaking them down into logical steps. Self-consistency methods \citep{Wang2022Self-Consistency}, further refine reasoning by aggregating multiple solutions to improve accuracy. Recent techniques like like \textit{tree of thoughts}, \textit{visualization of thought} \citep{Wang2023Review}, and \textit{program-aided language models} \citep{Gao2022PAL}  provide structured, interpretable reasoning pathways.

These prompting strategies demonstrate the potential for LLMs to handle spatial reasoning, but they also highlight the need for more robust methods to address limitations in multi-hop inference and natural language ambiguity. Our approach integrates prompting strategies with symbolic reasoning to enhance LLMs’ spatial reasoning capabilities more comprehensively.

\subsection{Effective Neural-Symbolic Integration}

Despite the success of deep learning, its limitations in reasoning, generalization and interpretability have spurred interest in hybrid approaches \citep{garcez2023neurosymbolic, hamilton2022neural}. Neural-symbolic integration aims to combine the strengths of neural networks (pattern recognition and uncertainty handling) with symbolic systems (logical reasoning and knowledge representation). Based on how neural and symbolic components interact and complement each other, research\citep{wanCognitiveAISystems2024} has identified key integration patterns: enhancing symbolic systems with neural capabilities (\textbf{Symbolic[Neural]}), combining neural learning with symbolic inference (\textbf{Neural|Symbolic}), compiling symbolic rules into neural structures (\textbf{Neural:Symbolic $\rightarrow$ Neural}), integrating logic rules through embeddings (\textbf{Neural $\otimes$ Symbolic}), and augmenting neural systems with symbolic reasoning (\textbf{Neural[Symbolic]}). These patterns can be grouped into practical architectures, including sequential, iterative, and embedded approaches, as well as LLM-based tool integration \citep{weber2019nlprolog, riegel2020logical, parisi2022talm}.

The emergence of LLMs has introduced new possibilities for neural-symbolic integration. Modern approaches leverage LLMs as powerful semantic parsers and natural language interfaces, while delegating the reasoning task to external off-the-shelf symbolic reasoners or solvers. The integration typically follows a two-step process: first, the LLM translates natural language queries into formal logical representations (e.g., Prolog predicates or ASP rules), then the logical solver performs structured reasoning to derive answers. However, this approach faces significant challenges in ensuring accurate and consistent mapping between natural language and logical forms. LLMs are not trained on logical programming language and they tend to generate logically inconsistent or syntactically incorrect programs, and the in-executable logical program will lead to the collpase of the whole neural symbolic system. For some complex and realistic dataset, the parsing  succesful rate can be as low as 17\% \citep{feng2024language}. To addrress this, researchers \citep{pan2023logic} have proposed self-refinement modules to refine the parsing results based on the solver’s erroneous messageses; however, the improvement of accuracy is not obvious with only 5\% after 3 iterations, highlighting the need for a more efficient architecture to enable sophisticated and robust feedback loop between LLMs and symbolic solvers.

This study leverages Dspy framework to construct a novel neural-symbolic pipeline that integrates LLMs with Answer Set Programming (ASP) in an iterative manner. Through its modular architecture and systematic optimization approach, DSPy can address many of the limitations faced by previous neural-symbolic integration attempts. By defining clear input-output signatures for each module in the pipeline, the framework ensures that the generated ASP code maintains consistent structure and enable the error feedback verification modules working in concert. Dspy's optimization compiler iteratively refines both the prompting strategies and weights used in each stage, ensuring consistent performance improvement, enabling more robust and adaptable neural-symbolic systems.

\subsection{Answer Set Programming}

Answer Set Programming (ASP) is a declarative programming paradigm that leverages predicate logic and stable model semantics to address complex combinational search tasks. Unlike procedural programming, which specifies step-by-step computations, ASP allows users to define solutions through logical relationships, enabling the solver to autonomously determine how to satisfy these conditions \citep{brewka2011answer}. ASP is case-sensitive, utilizing uppercase letters for variables and lowercase letters for constants, including atoms and predicates. The underscore character ``\_'' serves as a ``don't care'' variable that can be instantiated by any value. However, using placeholders in rule heads without corresponding definitions can lead to unsafe variable errors due to ambiguity during grounding.

ASP is closely related to Prolog, having evolved from it and sharing foundational concepts. Both languages utilize logical rules and facts for knowledge representation. However, while Prolog emphasizes procedural query evaluation, ASP focuses on generating answer sets —collections of ground atoms that satisfy the program's rules. This distinction allows ASP to handle nonmonotonic reasoning more effectively, accommodating multiple stable models for a single program, which is less common in Prolog.

The paradigm’s expressive power derives from four fundamental components, each serving distinct roles in knowledge representation and reasoning:

\textbf{Facts}:
Basic units of knowledge represented as unconditional logical atoms p(t1, ..., tn), where p is a predicate symbol and t1, ..., tn are terms. In spatial reasoning, facts typically represent object properties, locations, and basic spatial relationships.

\textbf{Rules}:
Logical implications of the form ``Head :- Body'', where the Head is an atom and the Body consists of literals. Rules enable inference, allowing new knowledge to be derived from existing facts.

\textbf{Constraints}: Special rules expressed as ``:- Body'' that eliminate answer sets violating specific conditions, thus serving as integrity checks within the knowledge base.

\textbf{Queries}: 
Goal-directed statements that extract information from the knowledge base, implemented as special predicates whose extensions yield the requested data.

In spatial reasoning contexts, ASP provides a robust framework for representing both static and dynamic spatial relationships through foundational predicates such as \texttt{block/1} for defining spatial regions, \texttt{object/5} for specifying object properties (identifier, size, color, shape, location), \texttt{is/3} for expressing spatial relationships, and \texttt{location/3} for precise coordinate definitions. This combination of predicates with ASP's logical foundations enables sophisticated reasoning about spatial configurations and relationships. 
	
	\section{Methods}
	
	\subsection{Datasets}
	
\textbf{StepGame}\citep{shi2022stepgame}

\texttt{StepGame} 
is a dataset designed for robust multi-hop spatial reasoning using a grid-based system, featuring eight directional spatial relations: top (north), down (south), left (west), right (east), top-left (north-west), top-right (north-east), down-left (south-west), and down-right (south-east). Each relation is defined by specific angles and distances for detailed visual representation, and it includes an ``overlap'' relation for scenarios where objects share the same location. The dataset supports reasoning hops from 1 to 10 across 10 subsets, each containing 10,000 samples corresponding to a single reasoning hop. However, the single Finding Relation question type may not fully capture the complexities of real-world spatial reasoning, as research indicates that large language models (LLMs) often struggle more with constructing object-linking chains from shuffled relations than with the spatial reasoning tasks themselves. Additionally, prior studies have identified template errors within StepGame that may skew model performance evaluations due to inadequate quality control during crowdsourcing, leading to inaccuracies in relationship mappings that can misrepresent an LLM's spatial reasoning abilities.

\textbf{SpartQA/SparTUN}\citep{mirzaee2022transfer}

SpartQA is built upon the NLVR (Natural Language for Visual Reasoning) images, featuring synthetically generated scenes depicting various spatial arrangements. Typically, each scenario consists of three blocks arranged either vertically or horizontally, with each block containing around four objects characterized by attributes like size, color, and shape. Besides, the dataset incorporates a wider range of spatial relationships, including 3D spatial reasoning, topological relations and distance relations. For FR questions, the candidate choices include ['left', 'right', 'above', 'below', 'near to', 'far from', 'touching', 'DK'], but there are more synonym relation names involved in the context and question.

SparQA differs significantly from StepGame in its complexity and scope. Its language structure is 2.5 times longer, featuring three blocks with approximately four objects each. The dataset handles multiple question types including Yes/No, FR, CO, and FB, with questions often involving complex relationships between multiple objects. For example, \textit{``What is the relation between the blue circle touching the top edge of block B and the small square?''} Except Yes/No questions, FR, CO and FB questions often have more than one answer. \texttt{SparQA} incorporates 3D spatial reasoning, topological relations, and distance relations, with options like \texttt{left}, \texttt{above}, and \texttt{near to}.

A distinctive feature of SparQA is its focus on  quantifier-based reasoning, with questions that test higher-level logic through universal (``all'') and exclusive (``only'') statements. For example, one third questions process queries like ``Are all of the squares in B?'' and ``Which block has only small black things inside?'' These questions require comprehensive evaluation of object sets and their properties rather than simple relational comparisons.

By utilizing these diverse datasets and question types, we aim to assess the spatial reasoning capabilities of LLMs across various complexity levels and spatial relation types.

	\subsection{LLM + ASP pipeline with DSPy}
	Recent studies have demonstrated LLMs' effectiveness as semantic parsers, often surpassing traditional parsing tools. While \citet{geibinger2023explainable} and \citet{eiter2022neural} showed promising results integrating LLMs with ASP, challenges remain. \citet{ishay2023leveraging} found LLMs could generate complex ASP programs but often with errors, while \citet{yang2023coupling} achieved 90\% accuracy on \texttt{StepGame} using \texttt{GPT-3} and \texttt{ASP}, their method's scalability remains uncertain.
	
	\begin{figure}[htp]
		\centering
		\fbox{\includegraphics[width=0.98\textwidth]{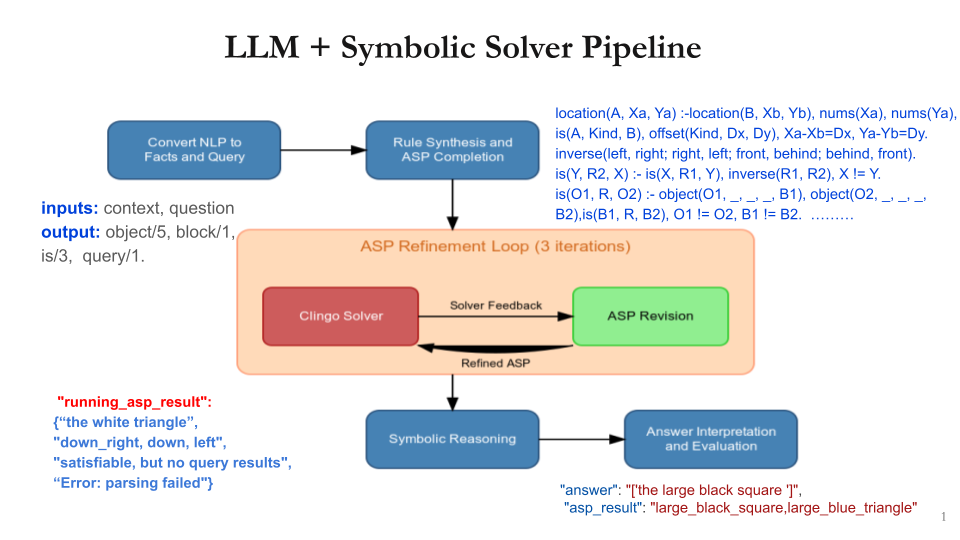}}
		\caption{LLM +ASP Pipeline  }
		\label{pipeline}
	\end{figure}
	
	Inspired by \citet{pan2023logic}'s LOGIC-LM framework and integration neural-symbolic strategies, we propose a novel neural-symbolic pipeline employing ASP using \texttt{DSPy} that treats the LLM as an agent capable of feedback and iteration. \texttt{DSPy}
    is a Python framework that uses a declarative and self-improving approach to simplify working with LLMs \citep{khattab2023dspy}. It automates the optimization of prompts and model tuning, enhancing reliability and scalability in AI applications. By defining tasks and metrics rather than manual prompts, \texttt{DSPy} streamlines the development of various NLP tasks and complex AI systems. The framework of this pipeline is shown in Fig.\ref{pipeline}.

	The pipeline consists of four main stages: \textbf{a) Facts Generation Stage}: LLM converts natural language descriptions into symbolic formulations and formal queries.
	 \textbf{b) ASP Refining Stage}: LLM iteratively refines the ASP representation over three iterations, adding rules, checking consistency, and incorporating feedback from error messages. \textbf{c) Symbolic Reasoning Stage}: The refined ASP undergoes inference using the Clingo solver, ensuring accurate and explainable reasoning by combining LLM capabilities with logical inference. \textbf{d) Result Interpretation and Evaluation}: This stage involves mapping the Clingo solver's outputs to candidate answers. For certain question types, like Yes/No and Finding-Block questions, the solver's output can directly serve as the correct answer. However, for \textit{Finding Relations} and \textit{Choose Object } questions, additional processing is necessary to filter relevant solutions. Outputs from the ASP solver are evaluated against a synonym dictionary to determine the accuracy. 
	
	 Overall, this pipeline requires multiple interactions with LLMs during ASP generation and refinement. We employ the DSPy framework to manage these complex workflows (e.g., interfacing with Llama3 60B DeepSeek and GPT 4.0 mini models via their APIs). DSPy's modular features enhance memory retention between modules, enabling adjustments and optimizations while maintaining workflow integrity. 
	
	Additionally, \texttt{DSPy} optimizes LLM prompts and weights, reducing the need for manual prompt engineering and ensuring consistent performance across datasets. Its optimization compiler iteratively generates and refines prompts, enhancing task performance. To support transparency and debugging, outputs from all modules are logged, capturing errors and providing insights for prompt engineering and system optimization, facilitating continuous improvement of the system and enhancing the integration of neural and symbolic components. In this way, this integrated neural-symbolic pipeline could greatly facilitate spatial reasoning in LLMs.

    
	To evaluate the effectiveness of our approaches in spatial reasoning comprehensively, we selected three representative LLMs with diverse architectures and capabilities: \texttt{DeepSeek}, \texttt{Llama3}, and \texttt{GPT4.0 Mini}. These LLMs were chosen to ensure a comprehensive assessment across different types of language representations, ranging from lightweight and specialized models like \texttt{DeepSeek} to more advanced general-purpose systems like \texttt{GPT4.0 Mini}. \texttt{Llama3}, known for its balance between performance and computational efficiency, provides an intermediate perspective. By testing our methods on these distinct models, we want to demonstrate the adaptability and robustness of our approach across a variety of LLM architectures and reasoning capacities. 

Given the task nature of multiple choices, we primarily employed exact match metrics for single-choice questions and partial match metrics for multiple-choice questions, aligning with the specific requirements of spatial reasoning tasks. To ensure accurate evaluation, we implemented custom post-processing to normalize responses and developed specialized metrics to handle both exact and partial matches between model outputs and ground truth.

\subsection{Baselines}
\textbf{Direct prompting}

As the most straightforward approach, it involves presenting the task to the LLM without additional guidance. The typical form is `` Given the question, please answer: While simple, this method is believed not to elicit the model’s full potential, particularly in complex reasoning tasks''. This straightforward approach involves presenting the task to the LLM without additional guidance. The basic structure consists of a simple template: ``Given the context and question, please answer the question by choosing from the choices''. While minimal, this method serves as a crucial baseline for model evaluation as it reflects the model’s fundamental ability to handle spatial tasks without any reasoning scaffolds. Despite its simplicity, this approach can achieve competitive results, particularly with well-trained models that have developed robust internal reasoning mechanisms.

\noindent\textbf{Facts + Rule prompting}

We explore an alternative approach that retains the benefits of structured knowledge representation while minimizing the complexity of formal logical programming. This method embeds logical rules directly into natural language prompts, testing LLMs’ ability to reason within structured frameworks.  Using the DSPy framework, this approach functions as a rule-based chain-of-thought (CoT) prompting strategy executed in two stages.

This approach aligns with the core principle of neural-symbolic AI: converting raw data into structured, symbolic representations for reasoning. By using predicates with precise argument structures, LLMs create consistent intermediate knowledge representations that facilitate question answering. This streamlined process maintains the advantages of formal reasoning while reducing computational complexity and implementation overhead.
	
In the LLM+ASP pipeline, generating a single ASP program requires multiple LLM calls due to the iterative refinement process, creating substantial computational overhead even with a modest three-iteration limit. To mitigate this, our alternative approach replaces formal ASP code generation with direct application of logical rules within natural language. The process begins by instructing LLMs to convert natural language inputs into structured facts using predefined predicates, a step retained from the LLM+ASP pipeline. Instead of formal logic programs, LLMs then apply relevant logical rules—such as inverse, symmetric, and transitive relations for the SparQA dataset, or offset-based chain-linking rules for StepGame—to derive answers through natural language reasoning.

By prompting LLMs to explicitly apply specific rules for different scenarios, this approach avoids dependence on external solvers. Consequently, it is termed “Facts+Rules.” This method offers a more straightforward and reliable path to spatial reasoning compared to generating and executing formal logic programs.



	\section{Experiment Results and Discussion}
	
In this section, we present the results of three methods: Direct Prompting Baseline, Facts + Rules Prompting, and Iterative LLM+ASP, on two benchmark datasets: StepGame and SparQA. We analyze their performance, computational complexity, and suitability for different types of spatial reasoning tasks. Additionally, we conduct an ablation study to assess the impact of iterative feedback loops in the LLM+ASP approach.

\subsection{StepGame}

\subsubsection{Implementation}

Another important aspect of the \texttt{StepGame} dataset is its reasoning hops range from 1 to 10 hops, distributed across 10 subsets. Specifically, each subset consists of 10,000 samples, each corresponding to a single reasoning hop. We sampled 300 instances for each reasoning hop 1 to 10, $k \in \{1, \ldots, 10\}$
, ensuring comprehensive evaluation across complexity levels. To standardize the task, we prompt LLMs to convert language descriptions into is\/3 and query\/2 facts using nine predefined spatial relations (e.g., left, right, top-right). Additional guidelines ensure accurate relation mapping, distinguishing between clockwise directions and cardinal references. This systematic approach maintains consistency and aids the models in handling spatial reasoning effectively. 

Our pipeline integrates a customized knowledge module adapted from \citet{yang2023neurasp}. This module employs coordinate-based reasoning rules, treating the second queried object as the reference point (0,0). By applying cardinal offsets, the module calculates the relative positions of connected objects, determining their spatial relationships iteratively. This method refines the models' intermediate inferences, enhancing accuracy in multi-hop tasks.

\begin{table}[ht]
\centering
\caption{Three Methods on StepGame (Accuracy \%)}
\label{tab:stepgame_accuracy}
\small 
\setlength{\tabcolsep}{4pt} 
\begin{tabular}{lccccccccccc}
\toprule
\textbf{Model \& Method} & \textbf{\rotatebox{90}{k=1}} & \textbf{\rotatebox{90}{k=2}} & \textbf{\rotatebox{90}{k=3}} & \textbf{\rotatebox{90}{k=4}} & \textbf{\rotatebox{90}{k=5}} & \textbf{\rotatebox{90}{k=6}} & \textbf{\rotatebox{90}{k=7}} & \textbf{\rotatebox{90}{k=8}} & \textbf{\rotatebox{90}{k=9}} & \textbf{\rotatebox{90}{k=10}} & \textbf{Overall} \\
\midrule
\multicolumn{12}{l}{\textbf{Deepseek}} \\
Direct       & 53.7 & 47.4 & 44.3 & 34.6 & 33.3 & 24.2 & 22.3 & 19.5 & 17.3 & 16.8 & 31.3 \\
Facts+Rules  & 61.7 & 58.2 & 55.5 & 48.3 & 46.2 & 42.7 & 38.6 & 35.2 & 32.5 & 30.8 & 44.9 \\
DSPy Pipeline & 93.7 & 89.2 & 92.5 & 89.3 & 88.5 & 87.7 & 86.3 & 85.2 & 84.5 & 79.8 & 87.7 \\
\midrule
\multicolumn{12}{l}{\textbf{Llama3}} \\
Direct       & 48.2 & 49.4 & 42.5 & 32.4 & 28.2 & 26.3 & 18.1 & 17.6 & 15.8 & 14.2 & 29.3 \\
Facts+Rules  & 58.2 & 56.4 & 54.5 & 50.2 & 48.9 & 45.4 & 42.8 & 40.1 & 38.3 & 35.7 & 47.0 \\
DSPy Pipeline & 82.2 & 76.4 & 79.5 & 81.2 & 77.9 & 80.4 & 72.8 & 70.1 & 69.3 & 65.7 & 75.6 \\
\midrule
\multicolumn{12}{l}{\textbf{GPT-4.0 mini}} \\
Direct       & 54.6 & 48.4 & 47.3 & 33.2 & 29.9 & 27.2 & 16.3 & 14.8 & 13.9 & 13.5 & 29.9 \\
Facts+Rules  & 62.8 & 60.4 & 58.3 & 52.3 & 48.6 & 45.5 & 42.2 & 38.3 & 35.5 & 32.9 & 47.6 \\
DSPy Pipeline & 85.8 & 82.4 & 85.5 & 85.3 & 87.1 & 85.5 & 79.2 & 75.3 & 72.5 & 68.9 & 80.8 \\
\bottomrule
\end{tabular}
\end{table}

\subsubsection{Results}
As shown in Table 1, the DSPy pipeline consistently outperforms baseline methods (Direct prompting and Facts+Rules) across all reasoning depths for Deepseek, Llama3, and GPT-4.0 mini models. This demonstrates the effectiveness of integrating linguistic processing with logical reasoning. For examples, at reasoning hop k=5, Deepseek’s accuracy jumps from 33.3\% (Direct) to 88.5\% (+55.2\%) using DSPy. Llama3 improves from 28.2\% to 77.9\% (+49.7\%), and GPT-4.0 mini increases from 29.9\% to 87.1\% (+57.2\%). These gains highlight the power of neural-symbolic feedback.

Accuracy declines as reasoning depth increases, reflecting task complexity. For example, at k=10, Direct prompting achieves only 16.8\% (Deepseek), 14.2\% (Llama3), and 13.5\% (GPT-4.0 mini), while DSPy maintains significantly higher performance: 79.8\%, 65.7\%, and 68.9\%, respectively.

Adding Facts+Rules boosts the overall accuracy to 44.9\%, 47.0\%, and 47.6\%, though it remains limited by sequential dependencies. In contrast, DSPy achieves the highest overall accuracy: 87.7\% (Deepseek), 75.6\% (Llama3), and 80.8\% (GPT-4.0 mini). It is obvious that  the DSPy pipeline’s iterative approach consistently enhances accuracy, particularly for deeper reasoning tasks, validating the benefits of combining symbolic reasoning with neural methods.

\subsubsection{Analysis}
Due to the controlled complexity and simple language structure, StepGame provides an ideal testbed for evaluating neural-symbolic integration for reasoning, allowing us to isolate the effects of multi-hop reasoning depth from linguistic complexity. There are no co-reference or named entity recognition issues since each agent is clearly defined (eg., ``A'', ``B'', ``C'', ``D''), and there is only one question type, querying the relation between two agents.

The success of LLM + ASP can be attributed to several key factors. The simplified predicate structure, utilizing only is/3 and query/2 predicates, provides a clear bridge between natural language and logical forms. This simplification, com- bined with our well-designed knowledge module, enables efficient handling of spatial relationships while maintaining robust error detection capabilities. 

Interestingly, the LLM+ASP method functions as a dataset quality checker. Aligning with \citet{yang2023neurasp}, we identify labeling errors in 10\% of the instances. Ambiguities in crowdsourced data accumulate with reasoning depth, revealing issues when models output multiple answers. This highlights the potential of neural-symbolic approaches not only for reasoning but also for improving dataset integrity.

\subsection{SparQA}

In order to evaluate the generalizability of our LLM+ASP approach against the more challenging benchmark SparQA, which focuses on complex natural language queries involving spatial relationships and requires precise fact extraction and reasoning over intricate descriptions. We aims to understand both the capabilities and limitations of neural-symbolic integration when confronted with realistic spatial reasoning tasks that more closely approximate real-world complexity. 

\subsubsection{Implementation}

 A representative test set of 220 examples, with 55 samples from each question type, was constructed to balance computational constraints and ensure comprehensive coverage. We also deliberately include challenging quantifier questions to assess the system’s capabilities. 
 
We adopted the same pipeline as \texttt{StepGame} to \texttt{SparQA}: (1) Converting Natural Language Context and Question to ASP Facts; (2) Adding Rules and Refining ASP Program; (3) Symbolic Reasoning; (4) Result Mapping and Evaluation. The first module prompted LLMs to identify blocks, objects, and relation facts using three predicates: \texttt{block/1}, \texttt{object/5}, and \texttt{is/3}, constraining the specific relation sets to minimize grounding errors. Rule refinement incorporates predefined inverse, transitive, and symmetric rules for spatial relations, enabling reasoning across multi-block environments. These rules were manually designed and updated, enhancing the system's capability to handle complex spatial queries that go beyond simple 2D reasoning tasks. The specific code and samples are seen in \textbf{Appendix}.

Key challenges in the implementation include parsing context-level descriptions for coreference resolution, representing implicit spatial relationships to avoid grounding errors, and managing object references with varying complexity (object/5). Additionally, query generation is difficult due to diverse question structures and complex quantification requirements. Despite careful prompt engineering, LLMs still struggle with generating queries that are different from the provided examples in the prompts. To overcome the challenge, we try to provide question type-wise query examples, quantification encoding query exampels to guide LLMs to write error free query. 
\subsubsection{Results}
    \begin{table}[ht]
    \centering
    \caption{Performance Comparison Across Methods on SparQA (Accuracy \%) }
    \label{tab:sparqa_performance}
    \begin{tabular}{l|c|c|c|c|c}
        \hline
        \textbf{Model \& Method} & \textbf{FR} & \textbf{FB} & \textbf{YN} & \textbf{CO} & \textbf{Overall} \\
        \hline
        \textbf{Deepseek} & & & & & \\
        Direct (Baseline) & 38.2 & 74.5 & 78.2 & 48.2 & 59.8 \\
        Facts+Rules       & 47.3 & 80.6 & 80.9 & 56.6 & 66.4 \\
        DSPy Pipeline     & 58.9 & 85.4 & 81.8 & 42.8 & 67.2 \\
        \hline
        \textbf{Llama3} & & & & & \\
        Direct (Baseline) & 26.9 & 65.7 & 72.5 & 55.8 & 55.2 \\
        Facts+Rules       & 46.8 & 70.2 & 81.4 & 57.1 & 63.9 \\
        DSPy Pipeline     & 53.1 & 83.3 & 80.5 & 60.8 & 69.4 \\
        \hline
        \textbf{GPT4.0 mini} & & & & & \\
        Direct (Baseline) & 45.4 & 60.9 & 58.2 & 57.6 & 55.5 \\
        Facts+Rules       & 54.3 & 68.2 & 50.4 & 61.5 & 58.6 \\
        DSPy Pipeline     & 65.3 & 80.5 & 64.8 & 72.7 & 70.3 \\
        \hline
    \end{tabular}
\end{table}

	As shown in Table~\ref{tab:sparqa_performance}, the neural-symbolic LLM+ASP pipeline showed mixed results across different models and question types. Finding Relation (FR) questions demonstrated significant improvement with accuracy increasing by approximately 20\% across all models (from 26.9\% to 53.1\% on Llama3, 38.2\% to 58.9\% on Deepseek, and 45.4\% to 65.3\% on GPT 4.0). Finding Block (FB) questions benefited from structured \texttt{block/5} predicate representation, showing substantial gains particularly in GPT 4.0 (from 60.9\% to 80.5\%). Choose Object (CO) questions showed varied results, with GPT-4.0 achieving a notable 15\% improvement while other models showed minimal changes. Interestingly, Yes/No (YN) questions does not benefit too much from complexity of neural-symbolic methods, only showing a slight improvement  overr with direct prompting across all models.

    SparQA include a lot quantifier reasoning question, almost one third questions include ``all, only, any''. For examples,``Which block has only squares inside?''. ``What block has all of the black objects inside of it?'', ``Are all of the triangles to the left of the black circle?'', Besides that, it also envolves a set, namely, using one attribute to represent the whole set of objects. Thanks to the object/5 atoms, we use use object(\_,\_,black,\_,\_)  in the query to match all the objects with all the black objects. Quantification typically challenges the pure neural approach. I If an LLM were to rely solely on natural language inference, it would need to exhaustively examine all objects with the black color. In contrast, converting the questions into logical expressions using the universal conditional (:) is relatively straightforward. As long as the logical expression is correctly encoded, the reasoning process becomes swift and accurate.

For example, ``What block has all the black objects inside of it?'' will be represented as: \begin{verbatim}
query(Block):- 
block(Block), not object(_, _, black, _, OtherBlock):
block(OtherBlock), OtherBlock != Block.
\end{verbatim}

As shown in the table, the Facts+Rules method achieves competitive and similar performance with the LLM+ASP approach while significantly outperforming direct prompting (more than 5\% improvement). While LLM+ASP achieves marginally higher accuracy in some cases, Facts+Rules offers advantages in implementation simplicity and reduced computational overhead. The improvement over baseline might be attributed to two factors: more consistent entity naming conventions in natural language prompts, and explicit instructions to follow the spatial logical rules.
    
\subsubsection{Analysis}
The error analysis revealed four primary categories of issues in the neural-symbolic system: parsing errors (31\%), grounding errors (23\%), satisfiable but no result issues (28\%), and wrong answers (18\%). Parsing errors, the most frequent issue, stemmed from syntax-related problems in ASP code generation, including unqualified relation specifications and improper predicate formatting. Grounding errors emerged from inconsistencies between variable naming and knowledge base content, while satisfiability issues arose when the provided facts and rules were insufficient for query resolution, particularly in cases involving implicit spatial relationships that the system failed to capture explicitly in the ASP code.

Each model exhibited distinct error patterns that impacted their performance differently. Deepseek demonstrated strength in handling complex spatial relationships but frequently generated syntax errors (42\% of its errors) and struggled with consistent comment handling in ASP code. GPT-4.0 mini showed promising initial code generation capabilities but consistently encountered issues with argument ordering in block/5 predicates, leading to fact-query inconsistencies. Meanwhile, Llama3 maintained consistent predicate formatting but showed higher rates of ``satifiable but no result'' failures, particularly in cases involving complex spatial relationship chains and nested relationships.
To address these challenges, the system would benefit from model-specific optimization strategies with tailored prompting approaches and custom validation rules, alongside an enhanced knowledge representation system capable of better handling implicit relationships and ambiguous spatial descriptions.	

\subsection{Impact of Feedback Loop in the DSPy-based Pipeline}

While language models demonstrate remarkable capabilities in semantic parsing task, their ability to generate executable logical programs remains challenging. Previous research points out that the direct translation from natural language to logical rules often results in low success rates, even below random baseline performance.\citep{feng2024language} This observation reinforces the importance of neural-symbolic integration especially the feedback loop between the two components, so that LLMs can benefit from symbolic solvers output, as shown in \citep{yang2023neurasp}.

To systematically analyze the effectiveness of our iterative feedback mechanism, we examine three primary error types in ASP solver execution: parsing errors (syntax errors, undefined predicates), grounding failures (infinite grounding scenarios, memory constraints), and solving stage failures (over-constrained conditions, inconsistent rules). Additionally, even successfully executed programs might produce solutions that don't align with ground truth labels due to semantic gaps between natural language specifications and logical encodings. Our feedback mechanism addresses these challenges through carefully designed prompts that instruct LLMs on error patterns and their respective fixes. It is hypothesized that the iterative feedback loop would reduce parsing errors and grounding failures.

	\begin{figure}[htp]
		\centering
		\fbox{\includegraphics[width=0.98\textwidth]{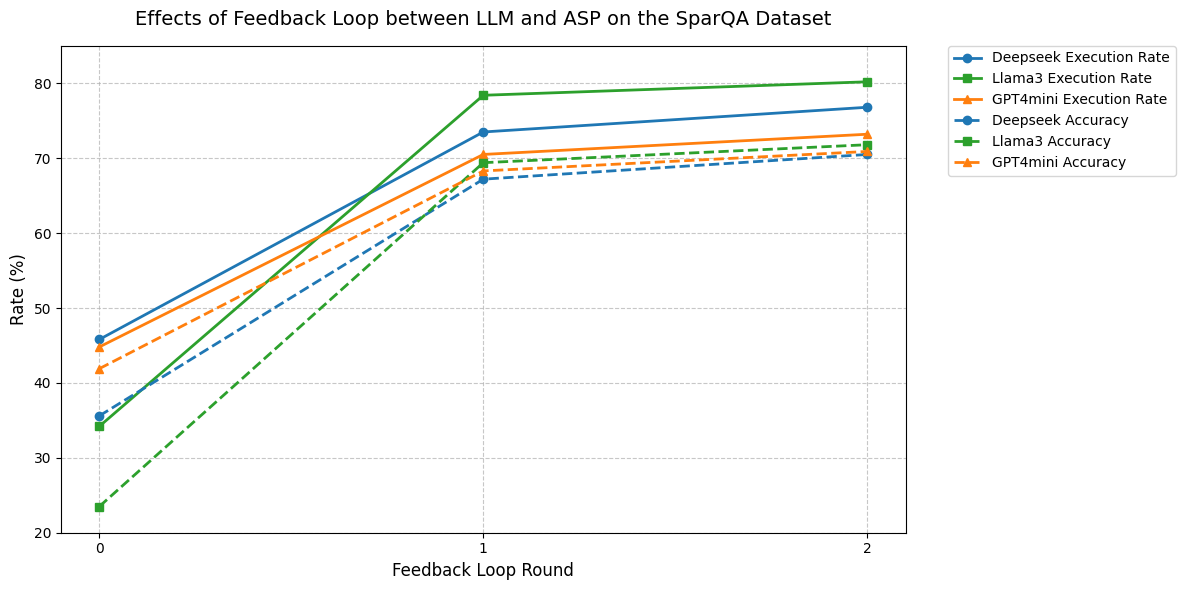}}
		\caption{Effects of Feedback Loop between LLM and ASP on the SparQA Dataset }
		\label{pipeline}
	\end{figure}
In the SparQA dataset, which features complex spatial reasoning tasks, our iterative feedback loop demonstrated substantial improvements across all models. As shown in Figure 2, the mechanism shows substantial improvements across all models: Deepseek's execution rate increased from 45.8\% to 76.8\%, Llama3 from 34.2\% to 80.2\%, and GPT4mini from 44.8\% to 73.2\% over two feedback rounds. Similarly, accuracy rates saw significant gains: Deepseek improved from 35.6\% to 70.5\%, Llama3 from 23.5\% to 71.8\%, and GPT4mini from 41.9\% to 70.9\%.

These results demonstrate that the feedback loop between LLMs and ASP solvers effectively addresses the inherent challenges in natural language to logic translation. The most substantial improvements occur in the first feedback round, with continued but diminishing gains in the second round. While additional feedback rounds might yield further improvements, we limited our experiment to three rounds due to computational costs. These findings demonstrate that the feedback loop significantly enhances both program executability and solution accuracy, establishing the effectiveness of our neural-symbolic integration approach.

\section {Conclusion}

This paper presents a neural-symbolic integration approach that significantly enhances LLMs' spatial reasoning capabilities. Our experimental results demonstrate that iterative feedback between LLMs and ASP solvers effectively improves both program executability and accuracy. The pipeline achieves an average 82\% accuracy on StepGame and  69\% on SparQA, marking substantial improvements over traditional approaches. For instance, our experiments demonstrate significant improvements over the baseline prompting methods, with accuracy increases of \textbf{40-50\%} on \texttt{StepGame} dataset and \textbf{ 8-15\%} on the more complex SparQA dataset. 

The success of neural symbolic integration stems from three key factors: (1) the effective separation of semantic parsing and logical reasoning, enabling precise control over each component; (2) the well-defined spatial relationships in a 2D environment, allowing for unambiguous predicate representation; and (3) the efficient handling of multi-hop reasoning chains through explicit logical rules.

We explored a simplified neural-symbolic approach, Facts+Rules, as an alternative to complex logical programming. This method showed modest improvements of 15-30\% over baseline prompting on both datasets and even a comparable perforamce as LLM+ASP pipleine on SparQA. This performance comparison reveals an important trade-off in neural-symbolic integration: LLM+ASP offers superior accuracy at the cost of implementation complexity and computational overhead, while Facts+Rules provides a more lightweight solution with reduced performance on structured tasks. These findings suggest that effective neural-symbolic integration can be achieved through different approaches, each offering distinct advantages in the balance between computational efficiency and reasoning capability.

The key contributions of our work include: (1) a systematic approach to boost spatial reasoning through neural-symbolic integration, (2) a cohesive pipeline that combines the strengths of LLMs and symbolic reasoning, and (3) robust knowledge representation techniques that generalize across different spatial reasoning tasks. Our iterative feedback mechanism particularly demonstrates the value of combining LLMs' natural language understanding with ASP's precise logical inference capabilities.

However, several limitations remain. The performance gap and the variable implementation complexity between StepGame (92\%) and SparQA (65\%) highlights the challenge of domain sensitivity in neural-symbolic systems. The pipeline struggles particularly with complex queries involving quantifiers and implicit spatial relationships. This study tackles the challenge by careful prompt engineering and providing more examples. Additionally, the conversion from natural language to logical programs remains error-prone, with execution rates varying significantly across different question types. Future work could finetune LLM on specialized logical program dataset and design more sophisticated feedback mechanisms and improved integration between neural and symbolic components.

This work establishes a foundation for enhancing LLMs' reasoning capabilities through neural-symbolic integration, driving forward the quest for more intelligent, interpretable, and efficient systems. In essence, the neural-symbolic approach treats LLMs as agents within a carefully orchestrated system. This perspective shifts the focus from improving individual LLM performance to optimizing the interplay between different components of the system. Future work should explore optimizing interactions between multiple models and reasoning components, involving more sophisticated orchestration techniques, improved integration of probabilistic reasoning with symbolic solvers, and employing the strengths of different LLMs at various stages of neural-symbolic systems.

	\vskip 0.2in
	\bibliographystyle{apalike}
	\bibliography{main}
	\newpage

		\end{document}